\documentclass[runningheads,a4paper]{llncs}
\usepackage{epsfig} 
\usepackage{epstopdf}
\usepackage{amssymb}
\usepackage{graphicx}
\usepackage{hyperref}
\usepackage{url}
\newcommand{\keywords}[1]{\par\addvspace\baselineskip
\noindent\keywordname\enspace\ignorespaces#1}

\begin{document}

\mainmatter  

\title{An Effective Image Feature Classification using an
improved SOM}

\titlerunning{An Effective Image Feature Classification using an
improved SOM}

%
%
\author{M. Abdelsamea \and Marghny H. Mohamed \and Mohamed Bamatraf}

\authorrunning{M. Abdelsamea et al.}

\institute{Assiut university, Assiut, Egypt
}

%
%

\tocauthor{Authors' Instructions}
\maketitle

\begin{abstract}
Image feature classification is a challenging problem in many computer vision applications, specifically, in the fields of remote sensing, image
analysis and pattern recognition. In this paper, a novel Self Organizing Map, termed improved SOM (\emph{iSOM}), is proposed with the aim of effectively classifying Mammographic images based on their texture feature representation. The main contribution of the \emph{iSOM} is to introduce a new node structure for the map representation and adopting a learning technique based on Kohonen \emph{SOM} accordingly. The main idea is to control, in an unsupervised fashion, the weight updating procedure depending on the class reliability of the node, during the weight update
time. Experiments held on a real Mammographic images. Results showed
high accuracy compared to classical \emph{SOM} and other state-of-art classifiers. 
\end{abstract}

\section{Introduction}

%
%

%
%
%
%
Image feature classification~\cite{CabralDCB14,CabralDCB11} presents a challenge in many computer vision applications. It plays a significant role in the fields of remote sensing, image analysis and pattern recognition. Recently, content-based image
classification and retrieval received increasing attention through
numerous applications \cite{Tommy2007} in the field of education, entertainment,
military and biomedicine. With the enormous growth of computational
power, image retrieval/classification have become more
demanding in the area of computer vision. However, the success of solving
such problems lies in the issues of object-based image
understanding, proper representation of image contents and suitable
learning algorithms.

The Self-Organizing Map (\emph{SOM}) \cite{Kohonen1990,kohonen2013} (also called Kohonen network) is an
artificial unsupervised network characterized by the fact that its
neighbouring neurons develop adaptively into specific detectors of
different vector patterns. The neurons become specifically tuned to
various classes of patterns through a competitive, unsupervised and
self organizing learning. The spatial location of a neuron in the
network (given by its coordinates) corresponds to a particular
input vector pattern. Similar input vectors correspond to the same
neuron or to neighbour neurons. One important characteristics of \emph{SOM}
is that it can simultaneously perform the feature extraction and it
performs the classification as well \cite{Neagoe2007}. 

In the medical field, \emph{SOM} has been used extensively in an efficient and effective way. In \cite{Santos2007} a classification
methods based on multilayer perceptrons and Kohonen self-organizing
map classifiers for image data to identify Alzheimer's
disease. Starting from the idea to consider the \emph{SOM} as a cell
characterizing a specific class only, Victor presents In \cite{Neagoe2002} a new
neural classification model called Concurrent Self-Organizing Maps
(\emph{CSOM}), representing a winner-takes-all collection of small \emph{SOM}
networks. Each \emph{SOM} of the system is trained individually to provide
best results for one class only. The \emph{CSOM} model proved to have better performances
than \emph{SOM}, both for the recognition rate and also for reduction of
the training time. In \cite{Tommy2007} Tommy proposes a new image
classification approach through a tree-structured feature set. In
this approach, the image content is organized in a two-level tree,
where the root node at the top level represents the whole image and
the child nodes at the bottom level represent the homogeneous
regions of the image. The tree-structured representation combines
both the global and the local features through the root and the
child nodes. The tree-structured feature data are then processed by
a two-level self-organizing map (\emph{SOM}), which consists of an
unsupervised \emph{SOM} for processing image regions and a supervising
concurrent \emph{SOM} (\emph{CSOM}) classifier for the overall classification of
images. DAR-REN et al. \cite{Chen2000405} applies the self-organizing maps (\emph{SOMs}) to classify the benign and malignant sonographic breast
lesions.

The classical \emph{SOM} and most of its variations rely on a fully unsupervised learning procedure. This is because of the node structure of the map does not provide any possibility of utilizing the class label while training the \emph{SOM} resulting in unstable behaviour when classifying pattern, especially in the real data with the presence of the noise and artifacts. Moreover, for the classification purpose, mapping can result in divided clusters because it requires that nearby points behave similarly. Motivated by the issues above, a simple but effective improvement in the classical \emph{SOM} is proposed in this paper aiming to integrate and take advantage of the probability of a particular node to be a winner by a voting criteria. 

The paper is organized as follows. Section~\ref{sec:SOMclassifier} provides a brief discussion on 
the advantages of using \emph{SOM} for the classification over the traditional classification models. In Section~\ref{sec:imagerepresentation}, an effective texture representation method is briefly discussed. Section~\ref{sec:iSOM} provides a discussion on the proposed model and its implementation. Then, an experimental study and conclusions with some possible future directions are provided in Sections \ref{sec:results} and \ref{sec:Conclusion}, receptively.

\section{\emph{SOM}-based Classifier}
\label{sec:SOMclassifier}

The \emph{SOM} neural network is one of the most popular
unsupervised neural network models, which simultaneously performs a
topology-preserving projection from the data space onto a regular
two-dimensional grid \cite{Kohonen1990}. There are some reasons to use a \emph{SOM} as a classifier: (i) Weights
representing the solution are found by iterative training, (ii) \emph{SOM}
has a simple structure for physical implementation and interpretation, (iii) \emph{SOM} can
easily map large and complex distributions and (iv) generalization
property of the \emph{SOM} produces appropriate results for the input
vectors that are not present in the training set \cite{Dokur2008951}. 

A basic \emph{SOM} network is composed of an input layer, an output layer,
and network connection layer. The input layer contains neurons for
each element in the input vector. The output layer consists of
neurons that are located on a regular, usually two-dimensional grid
and are fully connected with those at the input layer. The network
connection layer is formed by vectors, which are composed of weights
in the input and output layer.

The neurons in the map are connected to adjacent ones by a
neighbourhood relation dictating the topological structure of the
neurons. Each neuron $i$ in the map is
represented by an n-dimensional weight or reference vector
$w_{i}=[w_{1},\ldots,w_{n}]^{T}$, where $n$ is equal to the number
of neurons in the input layer.

When an input vector $x \in R^{n}$ is presented to the network, the
neurons in the map compete with each other to be the winner (or the
best-matching unit, \emph{BMU}) $b$, which is the closest to the input vector
in terms of some kind of dissimilarity measure such as Euclidean
distance as follows,
\begin{equation}\label{SOMDistnace}
    \|x-w_{b}\|=min\{\|x-w_{i}\|\}
\end{equation}

During training session, weights of neurons are topologically
arranged in the map within a certain geometric distance and are
moved toward the input $x$ using the self-organization learning rule
as represented in formula below :
\begin{equation}\label{SOMweight}
   w_{i}(t+1)=w_{i}(t)+\eta h_{bi}(t)[x(t)- w_{i}(t)]
\end{equation}
where $t=0,1,2,3,\ldots$ is the time lag, $\eta$ is a small positive
learning rate and $h_{bi}(t)$ is the neighborhood kernel around the
BMU $b$ at time $t$. In general, $h_{bi}(t)$ can be defined as
\begin{equation}\label{hci}
    h_{ci}(t)=h( \|r_{c}-r_{i}\| ,t)
\end{equation}
where $r_{c}, r_{i} \in R^{2}$ are the location vectors of neurons
$c$ and $i$, respectively, and when  $\|r_{c}-r_{i}\|$ increases,
$h_{ci}$ decreases to zero gradually. This leads to local relaxation
or smoothing effects on the weight vectors of neurons in the
neighbourhood of the \emph{BMU}. Therefore, similar input vectors are
grouped into a single neuron or neighbouring ones in the map when
learning is accomplished.

\emph{SOM} has been used as classifier by projecting the data vectors belong to
higher dimensional input space $n$ into $m$ many code-book vectors of
size $n$ organized in a two dimensional lattice structure. \emph{SOM}
provides two fundamental issues: the first is the clustering of data
and the second is the relationship between the clusters. The
clustering is an unsupervised learning while the relationship between
clusters can be seen in the planar surface by checking the distances
between the code-book vectors. Although it is difficult to deduce
exact relationship between those, since the code-book vector size is
much greater than the planar surface size of 2, this gives us an
insight about the classification regions. What is proposed here is totally
different from what was proposed in the previous works, a novel training algorithm for \emph{SOM} as well as a new node
structure to meet the proposed training is introduced. Another point is to apply the algorithm in a set of already selected features.

\section{Mammographic Image representation}
\label{sec:imagerepresentation}

In this section, a short discussion of an image texture feature extraction method is provide for effective Mammographic image representation.
 
The Grey Level Co-occurrence matrix (\emph{GLCM}) provides a full texture description of an image in a statistical fashion. Simply, the \emph{GLCM} technique computes first the probability of co-occurrence between two grey levels $i$ and $j$ given a relative orientation and distance for all possible co-occurring grey level pairs in an image window. Then, a set of selected statistics are
applied to the entire matrix to calculate the texture features. In this work, the four most commonly and practically used features (e.g., Dissimilarity, uniformity, entropy and contrast)~\cite{Haralick1973,sia2011,SOM2009} have been selected for the model evaluation. The following four statistics will be used exclusively in
this work:
\begin{enumerate}
    \item Dissimilarity  = $\sum_{i,j=1}^{G}C_{ij}|i-j|$.
    \item Uniformity = $\sum_{i,j=1}^{G}C_{ij}^{2}$.
    \item Entropy = $-\sum_{i,j=1}^{G}C_{ij}\lg C_{ij}$.
    \item Contrast = $\sum_{i,j=1}^{G}C_{ij}(i-j)^{2}$.
\end{enumerate}
Where, $C_{ij}$ represents co-occurring probabilities stored inside
\emph{GLCM}. $G$ represents number of grey level available.

\begin{figure*}[!htb]
\centering
\includegraphics[scale=.2]{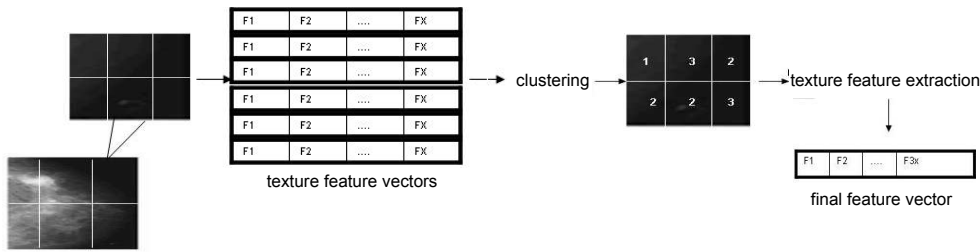}
\caption{The architecture of bloc wise feature extraction method with $SN=6$, $M=6$ and $L=3$ for simplification.}
\label{FigAg3}
\end{figure*}

For a more accurate feature extraction and a further investigation of the localization property of the represented features, the process of extracting textural information from Mammographic image depends on first identifying the object of interest as a reprocessing step. As a consequence, However, it is required to segment the images first as a pre-processing step before the feature extraction process. As a consequence, a bloc wise partitioning method \cite{Mohamed2008} is used in this work, which can be described as follows (see Figure \ref{FigAg3}):

\begin{enumerate}
	\item If the image contains inhomogenity regions then a set of $SN$ seeds are automatically selected and their associated regions are growing in a similar way to~\cite{sdr,abdelsamea20112d}. Otherwise, divide the entire image into $SN$ non-overlapping sub-images $SI=\{I_{1}, I_{2}, \ldots, I_{SN}\}$. 
	 \item Split each of these $SN$ sub-images into other $M$ blocks $I_{J}=\{B_{1},
B_{2}, \ldots, B_{M}\}$, $j=1, 2, \ldots, SN$.
    \item For each bloc $B_{i}, i=1,2,\ldots,M$, construct a bloc representing
set of texture feature vectors.
    \item Use the $k$-means algorithm to cluster the feature vectors into
several classes for each sub-image $I$ independently.
    \item For each cluster in $I_{i},i=1,2,\ldots,SN$, construct a sub-image
representing set of texture feature vectors $F_{K} =\{f_{1}, f_{2},
\ldots, f_{X}\}$, $k=1, 2, \ldots, L$; where $L$ is the number of
classes each of which contains $X$ texture features.
    \item Build the final set of texture features representing the overall
image in the form of a single transaction of the final dataset (set
of images),$T_{i}=\{t_{1}, t_{2}, \ldots, t_{c}\}$, where $c$ is the
number of images,  $t_{i}$ is a vector of the size ($SN \times L
\times X$), $i=1,2,\ldots,c$.
    \item For each $T_{i}=\{t_{1}, t_{2}, \ldots, t_{c}\}$ add the class label
of its image.
\end{enumerate}

\section{\emph{iSOM} Classifier}
\label{sec:iSOM}

As mentioned in previous sections, \emph{SOM} is designed to be unsupervised
learning technique. We here enhance the \emph{SOM} network to be used for
supervised learning (classification) by introducing a new node
structure and an enhanced learning utilizing the class label in the
weight update step.

First, every node is represented with the
set of connection weights $w=\{w_{0},w_{1},\ldots,w_{n}\}$ where $n$
is the number of attributes, and a set of winning class counters
($WCC^{m}$) $c^{m}=\{c_{1},c_{2},\ldots,c_{m}\}$ where $m$ is the
number of classes, this
representation provides the possibility of utilizing the class label
provided in the training set while training the \emph{SOM}. We can simply
say the vector $WCC$, is introduced to the node structure to provide
a voting criteria, so such nodes with maximum $WCC_{i}$ are pulled
during the weight update process. Shifting such nodes towards the
\emph{BMU} which is definitely of the same class increases the means of
relationship between such nodes. At the same time leaving nodes from
other class dims the relationship between such nodes and their
un-similar neighbours.

In every iteration after computing the distance between the input
vector and \emph{SOM} elements using \ref{SOMDistnace}, the winning node is
activated, this step is typically as proposed in classical Kohonen
\emph{SOM}, even though several distance function can be used, the main
idea is to measure the similarity between objects independently from
the data.

The next step after identifying the \emph{BMU} is to increase
the $WCC_{i}$ by one for the $i^{th}$ class accordingly. This
increment gives more confidence that this node is targeted by an
example of class $i$, this confidence indicates a similarity between
both the input example and the winning node.

At the final stage
what we call a constrained weight update is performed, the problem
with traditional \emph{SOM} is that all the neighbours are blindly attracted
or pulled towards the winning node, the term constrained means
selecting some node which are fitted to the criteria that is clear
here, those nodes which are mostly targeted by examples from the
same class are suppose to belong to the same cluster, and those
nodes not mostly targeted by examples from other classes should come
closer to this cluster, so they are left.

We can simply express
this as, when a winning node is activated, and before performing the
weight update, a vote is conducted, the only set of nodes in the
neighbourhood with maximum class counter that is equal to the current
instance class label will be considered as neighbour nodes. For an
instance $E_{j}$ belongs to class $C_{i}$, $X^{T} = Max \| WCC \|$,
where Max is a function that returns the set of nodes with maximum
$WCC_{i}$, finally the weight update given in \ref{SOMweight} will
be performed over $X$, at the same
the $WCC_{i}$ will be increased by one.

\begin{figure*}[!htb]
\centering
\includegraphics[scale=.2]{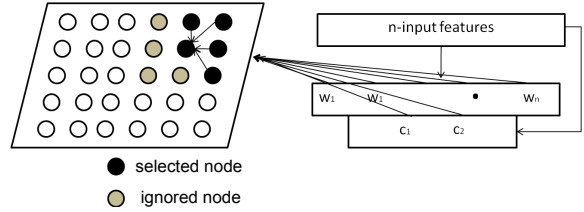}
\caption{The architecture of \emph{iSOM}: the proposed node structure and weight updating process.}
\label{FigAg4}
\end{figure*}

Figure \ref{FigAg4} illustrate the learning process of \emph{SOM} based on the proposed node structure. Every node is represented with the set of connection weights, and a set of winning class counters such that with maximum WCC (black nodes) are pulled during the weight update process. Shifting such nodes towards the BMU which is definitely of the same class increases the means of relationship between such nodes. At the same time leaving nodes (grey nodes) from other class dims the relationship between such nodes and their un-similar neighbours.

the rest of the algorithm will go according to the classical \emph{SOM}
explained the previous section, including the weight decrease given
in \ref{hci}, for relaxation.

Selecting nodes which are supposed
to be related to the winning node and update their weights towards
the winning node and leaving those nodes that are most probably
belong to different class, could enhance training in terms of time
as well as generated model quality.

\subsubsection{Implementation}

\begin{enumerate}
    \item Randomly set the initial values for all the connection
weights.
    \item For every node in the grid set the class counters to
zero.
    \item If training set is not empty, select an instance from
the training set and go to next step else go to step 7.
    \item Compute the distance between the selected instance and
every node in the network and select the wining node accordingly.
    \item For every node in the neighbourhood\\
    $              $- Select nodes where the maximum class counter is
the same as the winning node.\\
$              $- Update weights for the selected nodes in previous
step.\\
$              $- Increase the corresponding class counter of the node.\\
    \item Go to step 3.
    \item If no of epochs $>$ LIMIT then exit, Else go to step 2.
\end{enumerate}

\begin{table*}[!htb]
\begin{center}
\caption{The classification rate in terms of Precision, Recall and F-score by the \emph{iSOM} and \emph{SOM} models.}
\footnotesize
\label{table:2}
\begin{tabular}{c c  c  c c  c c  c}
\hline
\emph{SOM} map size &\multicolumn {3}{c} {\emph{iSOM} model} &\multicolumn {3}{c} {\emph{SOM} model} &\\
 & Precision& Recall & F-score & Precision & Recall & F-score \\
\hline
$10 \times 10$ &94.73 \hspace {5 mm}& 76.92 \hspace {5 mm}&84.9 \hspace {5 mm}& 80.39 \hspace {5 mm}& 79.12\hspace {5 mm}& 79.74\hspace {5 mm} \\
$15 \times 15$ &94.33 \hspace {5 mm}& 88.76 \hspace {5 mm}& 91.46 \hspace {5 mm}& 90.9 \hspace {5 mm}& 79.59\hspace {5 mm}& 84.86\hspace {5 mm} \\
$20 \times 20$ &97.87 \hspace {5 mm}& 85.26 \hspace {5 mm}& 91.13 \hspace {5 mm}& 98.18 \hspace {5 mm}& 93.1\hspace {5 mm}& 95.57\hspace {5 mm}\\
$25 \times 25$ &100 \hspace {5 mm}& 90.84 \hspace {5 mm}& 95.2 \hspace {5 mm}& 92.1 \hspace {5 mm}& 75.96\hspace {5 mm}& 83.25\hspace {5 mm} \\
\hline
\end{tabular}
\end{center}
\end{table*}

\section{Experimental Results}
\label{sec:results}
To demonstrate quantitatively the accuracy of the \emph{iSOM} model in the classification, we have used the Precision, Recall , and $F$-score metrics Such 
that 10-fold cross validation process is used. They are defined as follows:
\begin{equation}
{\rm Precision}   =  \frac{TP}{TP + FP}\,,
\end{equation}
\begin{equation}
{\rm Recall}   =  \frac{TP}{TP + FN}\,,
\end{equation}
\begin{equation}
F\hbox{-}{\rm score}  =  \frac{2PR}{P + R}\,,
\end{equation}
where $TP$, $FP$, and $FN$ represent, resp., the numbers of true positive (abnormal), false positive, and false negative (normal) foreground pixels. We also have compared the accuracy with \emph{SOM} model and other classifiers. In this experiment 142 images from the mini-MIAS database of mammograms data set \cite{1} are selected randomly for model
testing and evaluation. The sample is distributed between classes as
follows: Normal class (n=60), and abnormal class (n=82).

Table \ref{table:2} illustrates comparison between classical \emph{SOM} and enhanced
\emph{SOM} classifier to classify images based on image feature set
extracted using textural extraction method based on our Bloc-wise
\emph{ROI} selection method described in section \ref{sec:imagerepresentation} with $SN=6$, $M=8$ and $L=3$.

Comparison between the classical \emph{SOM} and \emph{iSOM} shows better
performance in terms of precision, recall as well as the
f-score with different map sizes. We can also say that the time consumed in training
both the networks differs and again the enhanced \emph{SOM} wins, as the
number of update operations performed by the enhanced \emph{SOM} is much
less compared to the classical \emph{SOM} due to the proposed constrained
weight update. One more point could be concluded from the experiment
indicates the enhanced \emph{SOM} could reach the best accuracy in less
time as we mentioned, as well as with less memory consumed if you
simply look to the map size. 

The proposed \emph{iSOM} couldn't only show
high accuracy when compared to the classical \emph{SOM} but comparing with
other famous classifiers in the field like Bayesian, Radial bases function network and others using weak
experimenter, see Table \ref{table:3}.

\begin{table}
\begin{center}
\caption{The classification accuracy of well-known classifiers.}
\footnotesize
\label{table:3}
\begin{tabular}{l|l|l|l}\hline
Classifier & Precision &\quad Recall &\quad F-score\\
\hline
 RBF & 58.76\% &\qquad 93.33\%&\qquad68.31\% \\
 Simple Logistic & 76\% &\qquad 76.08\%&\qquad76.05\% \\
 Bagging &87.50\% &\qquad 75.49\%&\qquad78.87\% \\
 J48 &93.10\% &\qquad 92.85\%&\qquad92.95\% \\
 NaiveBayes &58.97\% &\qquad 78.12\%&\qquad67.60\% \\
\hline
\end{tabular}
\end{center}
\end{table}


\section{ Conclusion and Future Work}
\label{sec:Conclusion}

In this paper, four texture features derived from the co-occurrence
matrix was used. For this, a
textural extraction method based on accurate \emph{ROI} selection for
obtaining efficient image representation has been utilized. 

The main characteristic of the \emph{SOM}-based classifier is the conservation of the topology: after learning, close observations are associated to the same class or to close classes according to the definition of the neighbourhood in the \emph{SOM} network. This feature allows considering the resulting classification as a good starting point for further developments. The paper presents a novel classifier inspired by the classical \emph{SOM}, termed improved \emph{SOM} (\emph{iSOM}), by introducing a new node structure and adopting the underlying self organizing learning procedure such that for the same number of
neurons, \emph{iSOM} has better recognition performances
than \emph{SOM}. Experimental results confirm the good
performance of the \emph{iSOM} when compared to other state-of-art classifiers.

As a future development, There are several research directions such as: 1) consider several and more effective image-based features (e.g., global information and other kind of local information), more samples for the evaluation and more  prototype-based algorithms for the comparison; 2) develop \emph{iSOM} to be working on a parallel fashion by applying more than 2 \emph{iSOMs} in a concurrent fashion and cope with the multi-class classification problems; 3) besides the use of more than 2 \emph{iSOMs}, we mention the possibility of extending the proposed model such that the underlying neurons are incrementally added/removed and trained to overcome the
limitation of manually adapting the topology of the network.

\bibliographystyle{elsarticle-num}
\bibliography{references.bbl}

\begin{thebibliography}{10}
\expandafter\ifx\csname url\endcsname\relax
  \def\url#1{\texttt{#1}}\fi
\expandafter\ifx\csname urlprefix\endcsname\relax\def\urlprefix{URL }\fi
\expandafter\ifx\csname href\endcsname\relax
  \def\href#1#2{#2} \def\path#1{#1}\fi

\bibitem{CabralDCB14}
R.~S. Cabral, F.~{De la Torre}, J.~P. Costeira, A.~Bernardino, Matrix
  completion for weakly-supervised multi-label image classification, IEEE
  Transactions Pattern Analysis and Machine Intelligence (PAMI).

\bibitem{CabralDCB11}
R.~S. Cabral, F.~{De la Torre}, J.~P. Costeira, A.~Bernardino, Matrix
  completion for multi-label image classification, in: Advances in Neural
  Information Processing Systems (NIPS), 2011.

\bibitem{Tommy2007}
T.~W. Chow, M.~K.~M. Rahman, {A new image classification technique using
  tree-structured regional features} 70~(4-6) (2007) 1040--1050.

\bibitem{Kohonen1990}
T.~Kohonen, The self-organizing map, Proceedings of the IEEE 78~(9) (1990)
  1464--1480.

\bibitem{kohonen2013}
T.~Kohonen, Essentials of the self-organizing map, Neural Networks 37 (2013)
  52--65.

\bibitem{Neagoe2007}
V.~Neagoe, A.-C. Mugioiu, C.-T. Tudoran, Concurrent self-organizing maps for
  multispectral facial image recognition, in: Computational Intelligence in
  Image and Signal Processing, 2007. CIISP 2007. IEEE Symposium on, 2007, pp.
  330--335.

\bibitem{Santos2007}
W.~Santos, R.~Souza, A.~e~Silva, P.~Santos~Filho, Evaluation of alzheimer's
  disease by analysis of mr images using multilayer perceptrons, polynomial
  nets and kohonen lvq classifiers, in: A.~Gagalowicz, W.~Philips (Eds.),
  Computer Vision/Computer Graphics Collaboration Techniques, Vol. 4418 of
  Lecture Notes in Computer Science, Springer Berlin Heidelberg, 2007, pp.
  12--22.

\bibitem{Neagoe2002}
V.~Neagoe, A.-D. Ropot, Concurrent self-organizing maps for pattern
  classification, in: Cognitive Informatics, 2002. Proceedings. First IEEE
  International Conference on, 2002, pp. 304--312.

\bibitem{Chen2000405}
D.-R. Chen, R.-F. Chang, Y.-L. Huang, Breast cancer diagnosis using
  self-organizing map for sonography, Ultrasound in Medicine and Biology 26~(3)
  (2000) 405 -- 411.

\bibitem{Dokur2008951}
Z.~Dokur, T.~Ölmez, Heart sound classification using wavelet transform and
  incremental self-organizing map, Digital Signal Processing 18~(6) (2008) 951
  -- 959.

\bibitem{Haralick1973}
R.~Haralick, K.~Shanmugam, I.~Dinstein, Textural features for image
  classification, Systems, Man and Cybernetics, IEEE Transactions on SMC-3~(6).

\bibitem{sia2011}
M.~M. Abdelsamea, Unsupervised parallel extraction based texture for efficient
  image representation, in: In Proceedings of the International Conference on
  Signal, Image Processing and Applications, SIA 2011, Chennai, India, 2011,
  pp. 6--10.

\bibitem{SOM2009}
M.~H. Mohamed, M.~M. Abdelsamea, Self organization map based texture feature
  extraction for efficient medical image categorization, in: In Proceedings of
  the 4th ACM International Conference on Intelligent Computing and Information
  Systems, ICICIS 2009, Cairo, Egypt, 2009.

\bibitem{Mohamed2008}
M.~Mohamed, M.~M. Abdelsamea, An efficient clustering based texture feature
  extraction for medical image, in: Computer and Information Technology, 2008.
  ICCIT 2008. 11th International Conference on, 2008, pp. 88--93.

\bibitem{sdr}
M.~M. Abdelsamea, An automatic seeded region growing for 2d biomedical image
  segmentation, in: In Proceedings of the International Conference on
  Environment and BioScience, Singapore, 2011, pp. 1--6.

\bibitem{abdelsamea20112d}
M.~Abdelsamea, An enhancement neighborhood connected segmentation for
  2d-cellular image, International Journal of Bioscience, Biochemistry and
  Bioinformatics 1~(4).

\bibitem{1}
\url{http://peipa.essex.ac.uk/info/mias.html}.

\end{thebibliography}

\end{document}